\documentclass[10pt, a4paper]{article}
\usepackage{booktabs}
\usepackage{multirow}
\usepackage[final]{lrec2026} 

\title{Headlines You Won't Forget:\\
Can Pronoun Insertion Increase Memorability?}

\name{Selina Meyer$^1$ \quad\quad~ Magdalena Abel$^2$ \quad\quad~ Michael Roth$^1$}

\address{$^1$Natural Language Understanding Lab \quad\quad $^2$Cognitive Psychology Lab\\
University of Technology Nuremberg\\
         \{firstname.lastname\}@utn.de\\}

\abstract{
For news headlines to influence beliefs and drive action, relevant information needs to be retained and retrievable from memory. In this probing study we draw on experiment designs from cognitive psychology to examine how a specific linguistic feature, namely direct address through first- and second-person pronouns, affects memorability and to what extent it is feasible to use large language models for the targeted insertion of such a feature into existing text without changing its core meaning. Across three controlled memorization experiments with a total of 240 participants, yielding 7,680 unique memory judgments, we show that pronoun insertion has mixed effects on memorability. Exploratory analyses indicate that effects differ based on headline topic, how pronouns are inserted and their immediate contexts. Additional data and fine-grained analysis is needed to draw definitive conclusions on these mediating factors. We further show that automatic revisions by LLMs are not always appropriate: Crowdsourced evaluations find many of them to be lacking in content accuracy and emotion retention or resulting in unnatural writing style. We make our collected \href{https://zenodo.org/records/19254945}{data} available for future work.
 \\ \newline \Keywords{News Memorability, LLM-based Text Editing, Cognitive Psychology} }

\begin{document}

\maketitleabstract

\section{Introduction}

News research in NLP is often related to boosting engagement of news articles, as formalized by behavioural data such as article dwell time \cite{davoudi-etal-2019-content}, likes, retweets, quotes, or replies \cite{gopalakrishna-pillai-etal-2025-engagement, park2021present}. This includes work on generating or editing suitable headlines or social media posts and 
mitigating the impact of misinformation
\cite{Srba2024-rf}. 

However, far less is known about how users process and retain news content, an equally critical factor in shaping belief and behaviour. Memorability plays a key role here: what users remember can influence what they believe and share. This is especially relevant in the age of generative AI, which has the potential to accelerate the production and spread of persuasive, yet misleading content \cite{spitale2023ai, bashardoust2024comparing, garry2024large}. Cognitive psychology, particularly the illusory truth effect, suggests that repetition alone can enhance perceived truthfulness and increase the likelihood of information being shared \cite{Pennycook2018-yf, Vellani2023-cz}, highlighting the importance of understanding other factors, such as linguistic characteristics, that can shape memory. 
\begin{figure}
    \centering
    \includegraphics[width=\linewidth]{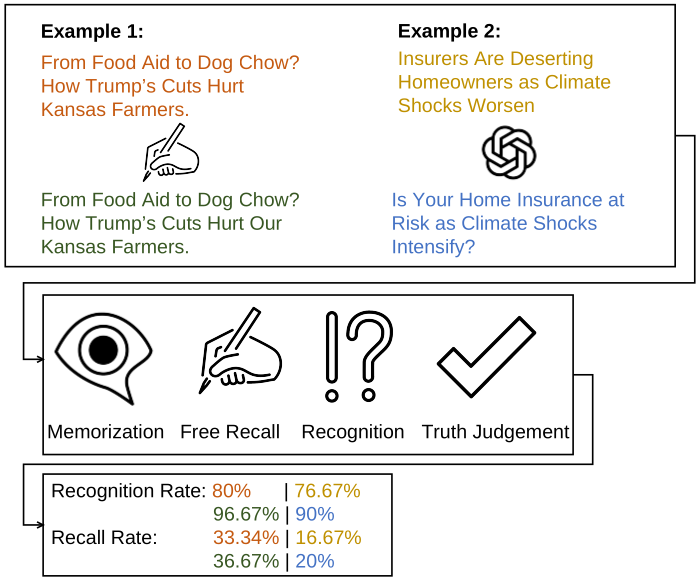}
    \caption{An overview of our experiment design. We ask humans and LLMs to insert first and second person pronouns in pre-existing news headlines. Participants are then shown headlines for a short time with the goal of memorizing them. In the examples shown,
    pronoun insertion considerably boosted
    recognition and recall.}
    \label{fig:abstract}
\end{figure}

While psychological drivers of belief in fake news have received considerable attention in cognitive psychology \citep[e.g.~see][]{pennycook2021psychology}, linguistic aspects that drive memorability of true or reputable news have largely been overlooked (cf.\ \S\ref{sec:relwork}). In this paper, we address this gap through a preliminary set of experiments that focus on the memorability of news headlines. In the course of this, we also explore LLMs' capabilities to manipulate news headlines, in the form of directly addressing readers, to make them more memorable. Our experiments 
test whether minor, targeted edits affect memory in terms of \textit{recognizing} and \textit{recalling} headlines, and whether LLMs can reliably implement relevant edits without distorting the original meaning (\S\ref{sec:methods}). Our results show that pronoun insertion has mixed effects on memorability  
and LLM revisions are not fully  
reliable~(\S\ref{sec:results}).

In summary, our contributions are two-fold: we test
LLMs
on a linguistically motivated paraphrasing task and we measure downstream effects in memorization studies using experimental methods from cognitive psychology.

\section{Related Work}
\label{sec:relwork}
Our work is related to text style transfer in that we manipulate one dimension of a text while preserving its core meaning \cite{mukherjee-etal-2024-large-language}, but differs in that we focus on a targeted manipulation
rather than changing the overall style of a piece of text. Prior research has shown that despite LLMs' generally impressive capabilities across many NLP tasks \cite{weiemergent}, even large LLMs often fail at simple tasks on which humans achieve perfect performance, such as writing sentences that contain a specific word, word unscrambling, or sentence editing 
\cite{efrat-etal-2023-lmentry, zhang-he-2024-large}. Fine-tuned models, even small ones, have been shown to outperform much larger base models on narrow text editing tasks, such as grammar correction \cite{raheja-etal-2023-coedit}, whereas zero- and few-shot prompting has been shown to lead to inconsistent performance in text style transfer tasks, including language detoxification and sentiment transfer \cite{mukherjee-etal-2024-large-language}, highlighting the continued importance of training data in such tasks. 

Specifically focusing on news rewriting and headline generation, \citet{gopalakrishna-pillai-etal-2025-engagement} explore different prompting strategies to rewrite news tweets to be more formal, casual, or factual, focusing on increasing predicted engagement, 
\citet{ao-etal-2021-pens} introduce a dataset of personalized headlines based on user preferences and candidate articles, and \citet{chen-etal-2023-honestbait} work on methods to leverage clickbaiting techniques, while 
keeping content faithful 
to increase 
reading interest and promote real information.

Beyond NLP-focused work, our approach is informed by findings from psychology, psycholinguistics and marketing showing that direct address and pronoun choice can influence memory even when propositional content is unchanged. For instance, \citet{symons1997self} discuss how information framed in relation to the self is more memorable, \citet{brunye2011better} show that second-person constructions induce stronger reader involvement, and \citet{cruz17consumer} suggest a robust effect of second person pronouns on consumer outcomes.
Related to news memorability, 
\citet{Lutz2024-mu} previously found different linguistic cues to affect 
cognitive and affective processing 
and \citet{Pena2023-fw} show that tweet-style texts are generally more memorable than news headlines. Also related to our work are studies by \citet{clark2026distinctive} and others on sentence recognition, which however do not take into account recall (i.e., the accessibility in memory in the absence of any retrieval cues). In contrast to this, we follow a common approach in cognitive psychology that provides a broader picture on memory by including measures for both recognition and recall \citep[e.g.][]{macleod1996word, unsworth2009examining}.

\section{Methods}
\label{sec:methods}
We performed a linguistic analysis on \citeauthor{Pena2023-fw}'s data. The findings suggest that personal pronouns help distinguish highly memorable content from less memorable items. To test whether this effect holds with headlines alone, we conducted a pilot study using topic-balanced headlines with and without pronouns. The results indicated that headlines with first and second person pronouns tend to be more memorable.\footnote{See Appendix \ref{sec:pena} for details on our analysis of \citeauthor{Pena2023-fw}'s data and pilot study.} 
Building on this insight, we explore the capabilities of a range of LLMs to insert such 
pronouns into real news headlines, without changing the content of the original headline or resulting in an unnatural writing style. Upon asserting the quality of the manipulated headlines, we conduct between-subject user studies to identify the effect of this specific linguistic change on memorability, in the absence of other discriminating factors. Overall, we run three memory studies, each informed by results of the preceding study. 

\subsection{Memory Studies}

Our memory study design is based on established study structures from the field of cognitive psychology \citep[e.g.~see][]{Pena2023-fw, abel2023item} and consists of five phases:

\begin{itemize}
    \item \textbf{Presentation Phase.} After reading and agreeing to the informed consent, participants view a fixed number of news headlines for 10 seconds each in random order, with no additional content shown. They are instructed to memorize them.
    \item \textbf{Distraction Phase.} Participants view and react to unrelated images for 60 seconds to reduce potential recency effects.
    \item \textbf{Recall Phase.} Participants freely recall and write down as many headlines as possible, aiming for exact wording. They are encouraged to spend at least 5 minutes on this task. If participants try to move to the next phase early, the system prompts them to take more time up to two times. After that, they may proceed even if less than 5 minutes have passed.
    \item \textbf{Recognition Phase.} All headlines of the presentation phase plus an equal number of unseen distractor headlines are shown in random order. For each headline, participants are asked to indicate whether they have seen the headline in the presentation phase.
    \item \textbf{Truth Judgement Phase.}  In addition to recognition and recall, we also measured perceived truthfulness. To this end, all headlines shown in the recognition phase are presented again in random order. Participants indicate how false or true they personally believe the headline to be on a 7-point likert-scale scale ranging from 
``definitely false'' to 
``definitely true''.  
\end{itemize}

\begin{table*}[htbp]
    \centering
    \resizebox{\textwidth}{!}{
    \begin{tabular}{p{0.5\textwidth}lcc}
    \toprule
    \textbf{Headline}&\textbf{Version}&\textbf{Recognition Rate}&\textbf{Recall Rate}\\
    \midrule
    \multicolumn{4}{l}{\textit{LLM Revisions which increased likelihood of recall and recognition}}\\
    \cmidrule{1-2}
         As the World Warms, Extreme Rain Is Becoming Even More Extreme&Original&73.34&26.67\\
         Are \textbf{You} Prepared for the Dramatic Increase in Extreme Rain as Earth Warms?&LLM Revision&86.67&46.67  \\
         As \textbf{Our} World Warms, Extreme Rain is Becoming Even More Extreme&Human revision&76.67&33.34\\
         \midrule
         Study finds no link between aluminum in vaccines and autism, asthma&Original&76.67&43.34\\
         Autism and Asthma: How a New Study Confirms No Connection to Aluminum in \textbf{Your} Vaccines&LLM Revision&90.00&43.34\\
         \midrule
         Insurers Are Deserting Homeowners as Climate Shocks Worsen&Original&80.00&20.00\\
         Is \textbf{Your} Home Insurance at Risk as Climate Shocks Intensify?&LLM Revision&86.67&30.00\\
         \toprule
         \multicolumn{4}{l}{\textit{LLM-Revisions which decreased likelihood of recall and recognition}}\\
         \cmidrule{1-2}
         From Food Aid to Dog Chow? How Trump’s Cuts Hurt Kansas Farmers.&Original&83.34&36.67\\
         \textbf{Your} Kansas Farmers Are Suffering: Trump’s Cuts Lead from Food Aid to Dog Chow&LLM Revision&63.34&26.67\\
         From Food Aid to Dog Chow? How Trump’s Cuts Hurt \textbf{Our} Kansas Farmers.&Human Revision&96.67&36.67\\
         \midrule
         Téa Leoni and Tim Daly Marry in Intimate New York Wedding&Original&83.34&46.67\\
         \textbf{Your} Inside Look at Téa Leoni and Tim Daly's Intimate New York Wedding&LLM Revision&80.00&33.34\\
         \midrule
         Kennedy Family Reunites for Massive Fourth of July Celebration&Original&90.00&36.67\\
         \textbf{You} Won't Believe the Kennedy Family's Massive Fourth of July Reunion!&LLM Revision&83.34&36.67\\
         \midrule
         \bottomrule
    \end{tabular}
    }
    \caption{Examples of headlines and LLM-revisions with their recognition and recall rates. Where available, equivalent human revisions are included for comparison.}
    \label{tab:examples}
\end{table*}
\paragraph{Study Material and Procedure} While all three studies share the same underlying design, they differ in study material, group assignments, and number of participants. We describe each study separately below. Across all experiments, each participant group has 30 participants, resulting in a total of 240 participants. The corresponding headline revision procedures are described in §\ref{sec:llm}.

 \paragraph{Study I}
    We collect 32 headlines from 32 major news outlets, with eight headlines for each of four topics: entertainment, politics, environment, and health, excluding those that originally include pronouns. Using various LLMs, we insert at least one first- or second-person pronoun into 16 headlines, collecting quality judgements by 8 annotators for each revision. We only include LLM revisions judged as both accurate and appropriate by at least 62.5\% of annotators.

    Participants are randomly assigned to one of two groups. In the presentation phase, each group sees 16 headlines, balanced across topics: 8 original and 8 LLM-revised with pronouns inserted. The assignment is counterbalanced: Group A sees one set revised and the other original, while Group B sees the reverse. 
    In the recognition phase, participants view the 16 headlines they have previously seen plus 16 held-out new ones which serve as distractor items (identical across groups). Of these new headlines, 7 are LLM-revised, ensuring that revised items are not recognized simply due to being the only ones containing pronouns. This design allows us to isolate the effect of pronoun use on memorability while controlling for content.  
\paragraph{Study II}
    Qualitative insights from study I results indicated that recognition improved when pronouns were organically integrated in the headline. We hypothesized that humans might achieve this more naturally than LLMs, which often relied on the addition of sentence fragments and clickbaity phrasing (see Table \ref{tab:examples} for examples). To investigate this, we include human revisions into our study material: in study II, half of the headlines with pronouns presented in the presentation phase are revised by prolific workers, while the other half is LLM-revised. Again, participants are randomly assigned to two counterbalanced groups. 

    \begin{figure}[htbp]
    \centering
    \includegraphics[width=\linewidth]{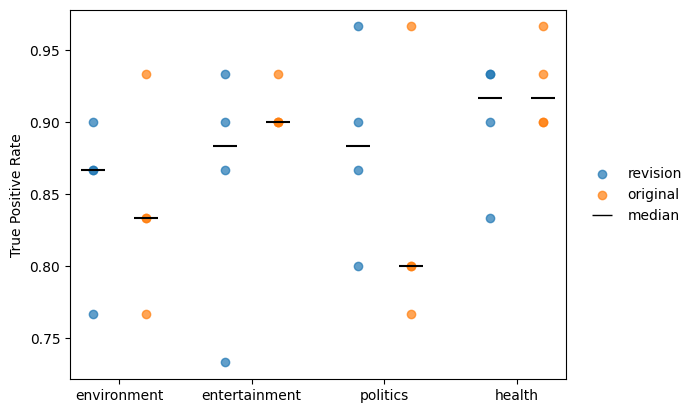}
    \caption{Mean true positive rates of original and revised headlines by topic (environment, entertainment, politics, health) in study II.}
    \label{fig:topic_effects}
\end{figure}

    \paragraph{Study III} Based on results obtained in study II, which indicate strong differences between effects of pronoun insertion across headline topics (see Figure \ref{fig:topic_effects}), we narrow down the selection of headlines to only one topic for study III. The differences between original headlines and revisions seemed to be strongest for headlines related to politics in study II, leading us to collect 32 new headlines from this topic. All 32 headlines are paired with a revision with pronouns inserted. We only use human revisions for this study and all revisions are written by the same person. 

    Participants are assigned to four groups with counterbalanced headline sets across presentation and recognition phases. Groups 1 and 3 view opposite versions of original and revised headlines during presentation, while Groups 2 and 4 see the held-out distractor sets from Groups 1 and 3. This design doubles the amount of evaluated headlines and tests whether pronoun use increases false recognition of previously unseen headlines.

\subsection{Pronoun Insertion}
\label{sec:llm}
\paragraph{Models} We use 8 LLMs of various sizes, including open-weight and proprietary models, to introduce first or second person pronouns to the collected headlines: \texttt{gpt-\allowbreak4o-\allowbreak mini-\allowbreak2024-\allowbreak07-\allowbreak18}, \texttt{gpt-\allowbreak4o-\allowbreak2024-\allowbreak08-\allowbreak06}, \texttt{Llama-\allowbreak3.1-\allowbreak8B-\allowbreak Instruct}, \texttt{Mistral-\allowbreak7B-\allowbreak Instruct-\allowbreak v0.3}, \texttt{Mixtral-\allowbreak8x7B-\allowbreak Instruct-\allowbreak v0.1}, \texttt{Qwen3-\allowbreak32B} (thinking mode enabled), \texttt{DeepSeek-\allowbreak V3-\allowbreak0324}, and \texttt{DeepSeek-\allowbreak R1-\allowbreak0528}. We set the temperature to 0.3 to allow for a limited amount of creativity and pass the same prompt to each model (see Appendix \ref{sec:models}).

\paragraph{Human Evaluation} We collect annotations that assess the accuracy and stylistic appropriateness of revised headlines compared to the originals. Based on the multidimensional quality metrics framework \cite{burchardt-2013-multidimensional}, we classify accuracy errors as misrepresentations, additions, or omissions, and style issues as grammar errors, awkwardness, or inconsistency.\footnote{See Appendix \ref{sec:annotation} for instructions provided to annotators.} 128 annotators recruited through Prolific review each original assigned to them alongside a revision and mark it as inaccurate or inappropriate only if at least one subcategory applies. They can also note shifts in tone or emotion. Before participating in annotation, annotators are required to pass a qualification test consisting of four original-revision pairs. Each original-revision pair receives 8 annotations. Annotators see 2–3 revisions per model and never see two revisions for the same headline. To compute inter-annotator agreement (IAA), we calculate the average raw agreement across annotators and annotation groups, as well as the mean of Krippendorff’s $\alpha$ across annotator groups. Acceptance rates for accuracy, style, and emotion retention are determined by the proportion of annotators who rated the accuracy and style as acceptable and did not report any shift in emotion or tone relative to the original headline. We collect annotations for 232 original–revision pairs derived from 29 seed headlines, yielding 1,856 judgments.

\paragraph{Human Rewriting} In addition to generating LLM revisions, we ask 10 Prolific workers who work in journalism, copywriting, or creative writing, to revise the original headlines. Participants are provided with instructions that are slightly modified from the prompt given to the LLMs (see Appendix \ref{sec:revisions}) and must pass a shortened version of the qualification test used in the LLM revision annotations in order to take part. Each participant rewrites 15 headlines and may skip headlines they feel are not suitable for pronoun insertion. We receive between 7 and 10 revisions per original headline. If two or more participants insert a pronoun in a headline in the same way, it is included in  study II (8 headlines overall). 

For study III, revisions for all 32 original headlines are obtained from a graduate student enrolled at the university of this work, who is an English native speaker. The student received the same instructions as the prolific workers and revisions were checked for appropriateness and faithfulness to the original meaning by the first author of this work. 

\section{Results and Discussion}
\label{sec:results}
We begin by outlining results regarding different LLMs' performances at the pronoun insertion task defined above as judged by human annotators. After this, we elaborate on the results obtained across the three user studies and offer exploratory analyses to identify potential mediating effects between pronoun insertion and headline memorability.
\subsection{LLM Revisions}
\label{sec:llm_results}

As a result of annotation, we observe IAA scores for accuracy, style and emotion retention of $\alpha$ = 0.19 (60.32\% raw agreement), $\alpha$ = 0.08 (56.64\%) and $\alpha$ = $-$0.03 (55.33\%), respectively, indicating that shifts in emotion between original headlines and revisions are especially subjective 
or difficult to judge for human annotators. 

\paragraph{Differences between LLMs}
\begin{table}[t]
    \centering
    \resizebox{\columnwidth}{!}{    
    \begin{tabular}{@{}l@{~~}c@{~~}c@{~~}c@{}}
    \toprule
         \textbf{Model}&\textbf{Accuracy}&\textbf{Style}&\textbf{Emotion} \\
    \midrule
         GPT-4o&\textit{62.9$^{\pm24.4}$}&\textit{65.5$^{\pm23.3}$}& 59.1$^{\pm20.6}$\\
         DeepSeek-chat&53.9$^{\pm27.2}$&63.4$^{\pm19.2}$& \textit{61.2$^{\pm25.3}$}\\
         DeepSeek-reasoning&\textbf{65.5$^{\pm24.0}$}&\textbf{67.2$^{\pm21.8}$}&\textbf{62.1$^{\pm19.6}$}\\
         GPT-4o-mini&57.3$^{\pm27.9}$&50.0$^{\pm22.9}$& \textcolor{red}{52.6$^{\pm24.0}$}\\
         Mixtral&49.6$^{\pm23.7}$&61.6$^{\pm22.1}$ & 54.7$^{\pm24.4}$\\
         Qwen&\textcolor{red}{47.4$^{\pm27.2}$}&51.7$^{\pm22.6}$& 56.5$^{\pm20.2}$\\
         Llama3-8b&50.0$^{\pm24.5}$& \textcolor{red}{47.8$^{\pm25.5}$}&53.5$^{\pm23.8}$\\
         Mistral&61.2$^{\pm30.9}$&51.7$^{\pm22.3}$& 60.3$^{\pm23.2}$\\
    \bottomrule
    \end{tabular}}
    \caption{Mean acceptance rates in \% per model. Highest values per category are \textbf{bolded}, second highest values are \textit{italicized}, lowest values are displayed in \textcolor{red}{red}.}
    \label{tab:LLM-rates}
\end{table}

Headlines rewritten by DeepSeek-reasoning have the highest acceptance rates across all three annotation categories, whereas Qwen displays the worst performance on accuracy, Llama on style, and GPT-4o-mini on emotion retention (see Table~\ref{tab:LLM-rates}). 
While revisions by larger LLMs generally seem to garner higher acceptance rates, this advantage is not consistent. For instance, the average accuracy acceptance rate for Mistral revisions is 11.64 and 7.32 percentage points higher than for Mixtral and DeepSeek-chat, and only 1.72 percentage points shy of GPT-4o's acceptance rate. This mirrors existing research on LLM's failures 
at solving simple text editing tasks out-of-the-box \cite{efrat-etal-2023-lmentry, zhang-he-2024-large}.

For the memorization experiment, we select LLM revisions with minimum acceptance rates of 62.5\% (corresponding to at least 5 out of 8 annotators) for style and accuracy. Mean acceptance rates of selected revisions across experiments lie at 81.9\% for style, 80.6\% for accuracy and 65.13\% for emotion retention.
\begin{table*}[hbtp]
    \centering
    \resizebox{\textwidth}{!}{
    \begin{tabular}{p{0.5\linewidth}|p{0.5\linewidth}}
         \textbf{Original Headline}&\textbf{LLM Revision} \\\toprule
         \multicolumn{2}{l}{\textbf{Inappropriate Role Attribution}}\\\toprule
         Bishop of major Catholic diocese exempts parishioners from Mass amid ICE raids& \textit{Why I’m Exempting You} from Mass During ICE Raids\\\midrule
         A couple tried for 18 years to get pregnant. AI made it happen&\textit{You and your partner} tried for 18 years. AI helped \textit{you} achieve it.\\\toprule
         \multicolumn{2}{l}{\textbf{Hallucinated Details}}\\\toprule
         Earth is as far away from the sun as it ever gets. So why is it so hot?&You’re far from the sun \textit{in July}, but why does it still feel so hot?\\\midrule
         Bishop of major Catholic diocese exempts parishioners from Mass amid ICE raids&Amid ICE raids, the bishop says you can skip Mass \textit{for your safety}.\\\toprule
         \multicolumn{2}{l}{\textbf{Insertion of Evaluative Statements}}\\\toprule
         Earth is as far away from the sun as it ever gets. So why is it so hot?&\textit{I'm Baffled:} Earth Is At Its Greatest Distance From The Sun, But Why Am I Still So Hot?\\\midrule
        Study finds no link between aluminum in vaccines and autism, asthma&Y\textit{our concerns about aluminum in vaccines and autism/asthma are valid}—but new research finds no link.	\\\toprule
         \multicolumn{2}{l}{\textbf{Ommission of Details}}\\\toprule
         Measles cases surge to record high \textit{since disease was declared eliminated in the US}&Are You Aware That Measles Cases Have Hit a Record High?\\\midrule
         Cierra Leaves ‘Love Island USA’ Due to a ‘Personal Situation’ Amid \textit{Backlash Over Resurfaced Post}&Cierra Leaves ‘Love Island USA’: Here’s What Her ‘Personal Situation’ Means for Fans Like You\\\bottomrule
    \end{tabular}}
    \caption{Examples of commonly found error types in LLM revisions of news headlines. Indicators for each error category are \textit{italicized}.}
    \label{tab:error_types}
\end{table*}
\paragraph{Common Errors in LLM Revisions}
We qualitatively examine the LLM revisions with style and accuracy acceptance rates of 50\% or less to identify common error types. Examples for each identified error type are presented in Table \ref{tab:error_types}. Revisions with low acceptance rates commonly include forms of inappropriate role attribution, which incorrectly frame the reader or writer as an active participant in the headline content. Other common error types include the addition of hallucinated details not present in the original headline, the insertion of evaluative statements, which introduce an author stance not grounded in the original, and the omission of details from the original. In some cases, combinations of multiple error types are displayed at the same time.

\subsection{Memory Experiment}
As evaluation measures, we compute the true positive rate for each presented headline and the false positive rate for each distractor based on user inputs collected in the recognition phase. We additionally calculate recall rate as the frequency of a headline's appearance in free recall divided by its presentation frequency. 

\paragraph{Recall Matching} Recalled items are manually matched to their corresponding headlines and ambiguous cases (e.g., items matching multiple headlines) are not counted. For instance, all items containing the word \textit{NASA} and no reference to another headline were matched to the headline \textit{NASA Website Will Not Provide Previous National Climate Reports} or its revision, depending on experimental group. We provide some examples for recalled items and their respective original headlines in Table \ref{tab:recall_examples}.  This resulted in some items displaying a considerate amount of distortion or lack of detail compared to the headlines seen by participants in the presentation phase. To account for this, we measure recall distortion using the mean cosine similarity, based on S-BERT embeddings \cite{reimers-gurevych-2019-sentence}, between recalled items and their respective ground truth headline. High similarity with the original means that participants remembered a headline in detail and correctly, whereas low similarity is an indicator for a participant remembering merely the gist of a headline or even misremembering the content of the headline.

\begin{table}[t]
    \centering
    \resizebox{1.0\columnwidth}{!}{
    \begin{tabular}{@{}lrcrr@{}}
    \toprule
         \multicolumn{3}{r}{\textbf{Recognition TP / FP rates (\%)}} & 
         \textbf{Recall (\%) (cos. sim.)} \\
         \midrule
        \textit{Study I}&&&\\
         \cmidrule{1-1}
         \textbf{original}&85.4$^{\pm7.2}$&~~9.0$^{\pm2.5}$ & 34.2$^{\pm12.8}$ (80.2$^{\pm5.8}$)\\
         \textbf{revision}&84.0$^{\pm8.1}$ &~~8.1$^{\pm3.2}$ &32.9$^{\pm10.5}$ (80.5$^{\pm6.9}$)\\
        \midrule
        \textit{Study II}&&&\\
         \cmidrule{1-1}
         \textbf{original}&87.7$^{\pm6.7}$&~~9.4$^{\pm2.5}$& 34.4$^{\pm12.1}$ (79.1$^{\pm7.1}$)\\
         \textbf{revision}&87.3$^{\pm6.4}$ &10.5$^{\pm5.5}$ &34.2$^{\pm14.3}$ (75.4$^{\pm5.9}$)\\
         \midrule
        \textit{Study III}&&&\\
         \cmidrule{1-1}
         \textbf{original}&80.9$^{\pm8.6}$&~~6.2$^{\pm4.2}$ &30.5$^{\pm10.2}$ (76.6$^{\pm9.7}$)\\
         \textbf{revision}&80.4$^{\pm9.2}$ &~~7.2$^{\pm4.9}$ &28.9$^{\pm10.7}$ (76.0$^{\pm8.7}$)\\
         \bottomrule
    \end{tabular}}
    \caption{Mean rates of true positive (TP) hits and false alarms (FP) for \textit{recognition} and average \textit{recall} rates.}
    \label{tab:main}
\end{table}

\begin{table*}[tb]
    \centering
    \resizebox{\textwidth}{!}{
    \begin{tabular}{p{0.35\textwidth}p{0.45\textwidth}p{0.11\textwidth}}
    \toprule
         \textbf{Recalled Text}&\textbf{Presented Headline}&
         \textbf{Cosine Similarity} \\
         \midrule
         New study shows no link between aluminum in vaccines and autism, asthma& Study finds no link between aluminum in vaccines and autism, asthma&1.00\\
         \midrule
         Beyond the Lights: How Fireworks Affect you and your Animals&Beyond the Light Show: How Fireworks Affect You and Your Animals&0.99\\
         \midrule
         John Goodman shows 200 lb weight loss&John Goodman Shows You His 200-Lb. Weight Loss Transformation.&0.88\\
         \midrule
         JD Vance is pushing trumps agenda&How JD Vance shapes and sells the ‘Trump doctrine’ on foreign policy&0.72\\
         \midrule
         jd vance forign affairs&What You Need to Know About JD Vance Shaping the ‘Trump Doctrine’ on Foreign Policy&0.65\\
         \midrule
         Brad Pitt and Angelina Jolie&Brad Pitt Takes a Bold Step in His Legal Battle with Ex-Wife Angelina Jolie Over Their French Winery: Here's What You Need to Know&0.51\\
        \midrule
NASA will no longer monitor rain sometime&NASA Website Will Not Provide Previous National Climate Reports&0.47\\
         \midrule
         Trump economy. &From Food Aid to Dog Chow? How Trump’s Cuts Hurt Kansas Farmers.&0.31\\
         \bottomrule
    \end{tabular}
    }
    \caption{Examples for items recalled in the free recall phase and their cosine similarity with the corresponding headlines. While some recalled items are mostly faithful to the original headline shown during the presentation phase, others merely reproduce individual words or a general gist.}
    \label{tab:recall_examples}
\end{table*}

\begin{table}[htbp]
    \centering
    \resizebox{1.0\columnwidth}{!}{
    \begin{tabular}{lc}
    \toprule
        Study I&\\
        \cmidrule{1-1}
        Recognition TP&$t(30)=0.54, p=0.59$\\
        Recognition FP&$t(14)=0.66, p=0.52$\\
        Recall&$t(30)=0.3, p=0.76 $\\
        cos.-sim.&$t(30)=-0.13, p=0.9$\\
        \midrule
        Study II&\\
        \cmidrule{1-1}
        Recognition TP&$U=136.5, p=0.76$\\
        Recognition FP&$t(14)=-0.5, p=0.62$\\
        Recall&$t(30)=0.04, p=0.96 $\\
        cos.-sim.&$t(30)=1.57, p=0.13$\\
        \midrule
        Study III&\\
        \cmidrule{1-1}
        Recognition TP&$U=538, p=0.73$\\
        Recognition FP&$U=456.5, p=0.46$\\
        Recall&$t(62)=0.64, p=0.53 $\\
        cos.-sim.&$U=539, p=0.72$\\
         \bottomrule
    \end{tabular}}
    \caption{Results of statistical tests for main measures across the three studies. }
    \label{tab:main2}
\end{table}

\paragraph{Effects of Pronoun Insertion on Memorability} Results for all three studies are summarized in Table \ref{tab:main}. We run significance tests for the main measures on each study separately. We use two-tailed independent t-tests for normally distributed data and Mann-Whitney U tests were parametric assumptions are violated (see Table \ref{tab:main2}). 
Overall, the effects of pronoun insertion on the 
evaluation measures considered across the three studies are not significant, indicating that pronouns alone do not systematically affect news headline memorability. 

\paragraph{Effects of Pronoun Insertion on Perceived Truthfulness} Mean results for perceived truthfulness given headline version (original or revision) and whether it had been presented in the presentation phase or not are provided in Table \ref{tab:cred}. 
\begin{table}[htbp]
    \centering
    \resizebox{1.0\columnwidth}{!}{
    \begin{tabular}{@{}llccc@{}}
    \toprule
    &&\multicolumn{3}{c}{\textbf{Truth Judgement}}\\
         &\textbf{Seen}&\textit{Study I}&\textit{Study II}&\textit{Study III} \\
         \midrule
         \textbf{Original}&False&4.55$^{\pm1.6}$&4.42$^{\pm1.6}$&4.60$^{\pm1.7}$\\
          &True&5.22$^{\pm1.6}$&5.11$^{\pm1.7}$&4.81$^{\pm1.7}$\\
        \cmidrule{1-1}
         \textbf{Revised}&False&5.04$^{\pm1.5}$&4.76$^{\pm1.8}$&4.41$^{\pm1.8}$\\
              &True&5.26$^{\pm1.6}$&5.11$^{\pm1.7}$&4.74$^{\pm1.8}$\\
\bottomrule
    \end{tabular}}
    \caption{Mean truth judgements of original and pronoun-inserted presentation headlines (seen) and distractor items (unseen).}
    \label{tab:cred}
\end{table}
Although all original headlines included in the study were collected from reputable news venues (NYT, NPR, CNN, Yahoo news, CBS, CNBC, NBC, Washington Post, USA Today, and Forbes) and only revisions with a high accuracy acceptance rate were included in the study, we reasoned that the introduction of first and second person pronouns might affect truth judgements. Past work has also found evidence of the illusory truth effect: repetition affects perceived truthfulness of information \cite{Pennycook2018-yf, Vellani2023-cz}. 
Like for the main measures, no difference in truth judgements between original and revised headlines was found across studies. We do observe slight, though statistically insignificant indications of illusory truth effect, meaning that headlines encountered in the presentation phase were considered slightly more truthful on average than headlines first seen in the recognition phase. 

\begin{figure}
    \centering
    \includegraphics[width=\linewidth]{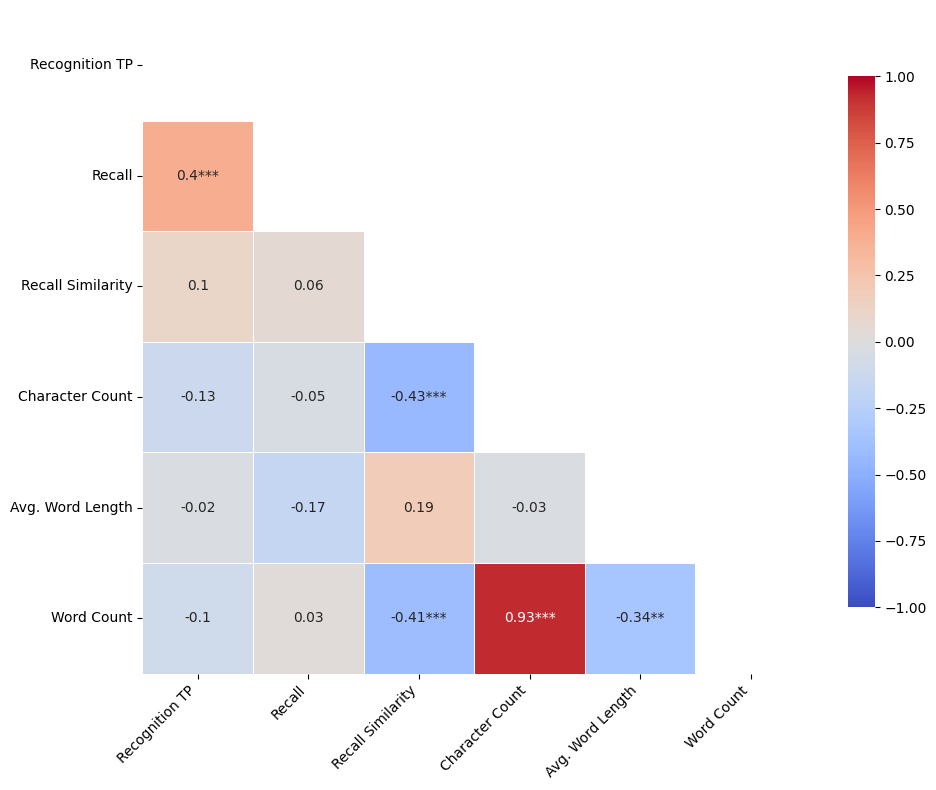}
    \caption{Bonferroni-corrected Pearson correlations between memory measures and headline length features. * denotes significant correlations at p$<$0.05, ** at p$<$0.01, and *** at p$<$0.001.}
    \label{fig:corrs}
\end{figure}

\paragraph{Exploratory Analysis and Interpretation}
On average, revised headlines were longer and had shorter words (average character count: 81.09$^{\pm21.37}$, average word length: 4.84$^{\pm0.58}$) compared to original headlines (67.78$^{\pm15.07}$, 5.18$^{\pm0.7}$). To identify to what extent this might impact memorability, we pool all headlines across the three studies and perform a correlation analysis, taking into account recognition and recall rates, cosine similarities of recalled items, headline lengths, and average word lengths (see Figure \ref{fig:corrs}). We find a significant negative correlation between recall similarity and headline length based on both word and character counts, but recall and recognition rates show no clear correlation with length features. This indicates that longer headlines tend to be remembered in less detail, but are retrievable and recognizable at similar rates as shorter headlines. We also find a significant positive correlation between recognition and recall rates, indicating that 
if a headline can be recognized correctly, it also tends to be accessible in the absence of any retrieval cues (and vice versa).

Qualitative inspection of recognition and recall rates further reveals individual headlines for which the addition of a pronoun clearly increases or decreases recognition. As shown in the examples provided in Table \ref{tab:examples}, it becomes apparent that revisions that naturally incorporate pronouns—either through restructuring or simple pronoun insertion—show a tendency to improve memorability, while clickbaity or unnatural edits seem to reduce it. 
\begin{figure}[t]
    \centering
    \includegraphics[width=\linewidth]{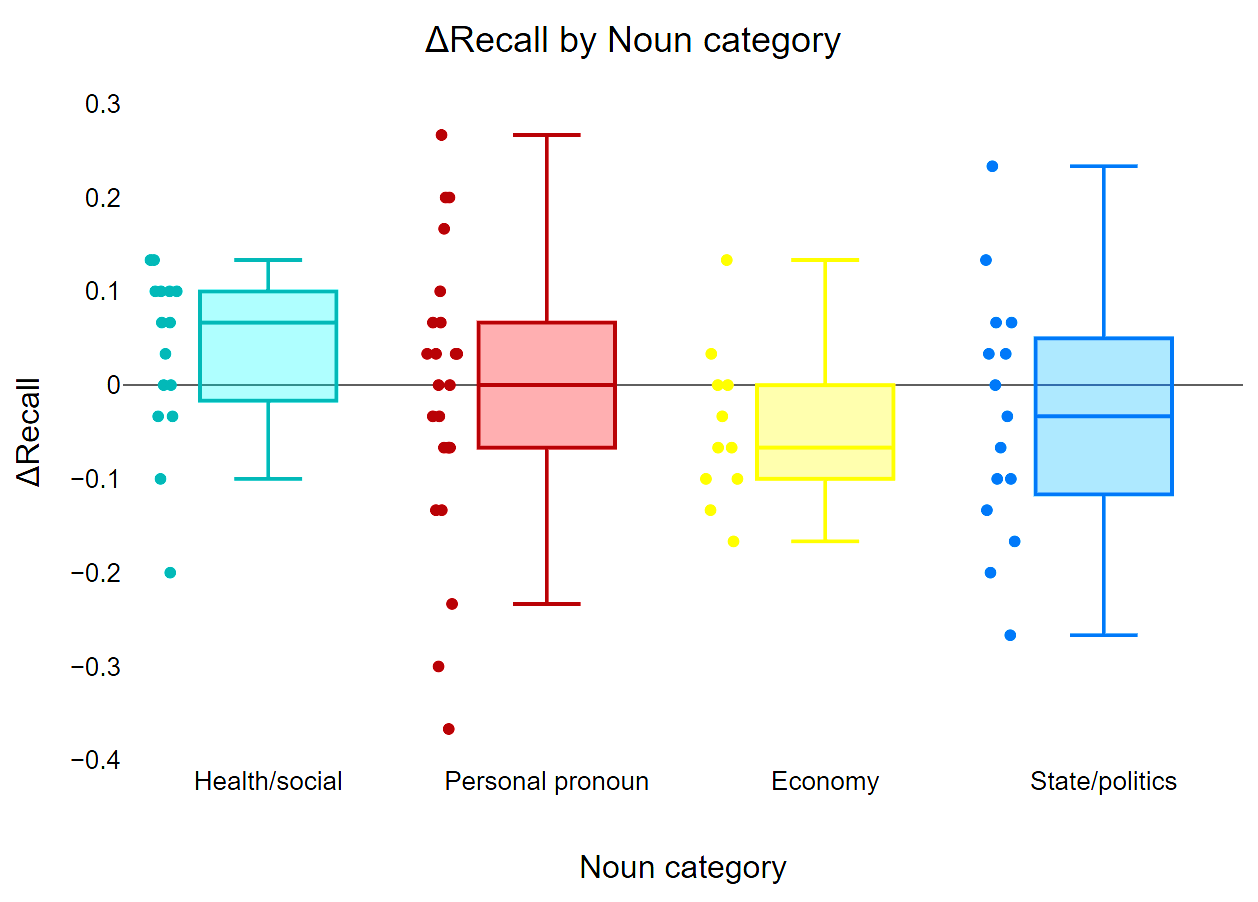}
    \caption{Increases/decreases in recall rate between original and revised headlines for insertions by nominal category, i.e.~personal pronouns (``we'', ``you'') and possessives with different nominal heads: Health/social (e.g.~``your insurance'', ``our babies''), Economy (e.g.~``your dollars'', ``our farmers'') and State/politics (e.g.~``your country'', ``our election'').}
    \label{fig:pron_contexts}
\end{figure}

Moreover, effects on recall are 
influenced  by the immediate context of the inserted pronoun: in particular, possessive pronouns with social/health-related nominal heads seem to improve recall, but economic ones do not (e.g.~``our babies'' vs.~``our farmers'', see
Figure \ref{fig:pron_contexts}). A potential reason for this might be that information related to health and social factors has a higher potential to be personally relevant and contain actionable information, compared to political and economic information which is less directly controllable by consumers of news. Future work could incorporate judgements of personal relevance to verify this.

\section{Conclusion}
In this paper, we present computational experiments targeting a specific linguistic change, namely the insertion of first- and second-person pronouns into news headlines, along with user studies examining their effects on memorability. We show that LLMs do not always insert pronouns appropriately, as indicated by crowdsourcing evaluations. Collectively, our memory studies lead to the conclusion that pronoun insertion in itself has no consistent effect on memorability. 
A closer look revealed individual cases of memorability impairment and enhancement, with substantial variation
across 
contexts 
suggesting a need for more fine-grained analyses and additional data. Moving forward, we plan to investigate other linguistic features that may more strongly influence memorability and to expand our evaluation of LLM capabilities in this context. 
Ultimately, our goal is to develop computational methods for making news and other information encountered online more memorable while preserving its original content. We believe that this approach has the potential to boost true, high-quality information over  misinformation, thus complementing other efforts towards the mitigation of the impact of misinformation on society. To support further research, we release our data, including revision annotations and 7,680 unique memory and truth judgments from three experiments involving 240 participants.\footnote{\url{https://zenodo.org/records/19254945}} We encourage NLP researchers to consider memorability as a modelling feature alongside engagement and related factors.

\section{Limitations}
We identify several limitations in this work, which we describe below: 

We are aware of potential interaction effects of other aspects related to pronoun insertion, such as increases in headline length or long words, changes in syntactic structure, or use of loaded language, which might be responsible for some headlines gaining a boost in memorability, while others saw deterioration when pronouns were inserted. In the future, we plan to develop computational methods to make targeted changes, such as the ones presented, while changing as little as possible of the remaining text. We also plan to explore more impactful changes to news texts, including headlines, to explore a greater variety of linguistic features which can impact memorability in this context, and run studies on larger scales, including more study items in the process, to increase the robustness of experimental results. While we are confident that the set of experiments presented here is suitable to conclude that pronoun insertion alone does not consistently boost memorability of news headlines, we are aware that the comparably small number of included headline items limits our ability to explain why some headlines benefit from pronoun insertion, whereas for others it leads to a decrease in memorability. Consequently, in future studies, we will focus on creating a database large enough to uncover linguistic patterns that interact with changes to increase or decrease memorability.

In cognitive psychology, experiments as the one described in this paper are often conducted in lab settings. While using Prolific for data collection yields many advantages, it also increases the likelihood of participants using external tools to help them remember headlines or be exposed to external distractors. To counteract this, we specifically instructed participants not to use external tools and make sure they are undisturbed for the duration of the study. We also embedded the news headlines as images instead of text, to prevent participants from copy-pasting content for later study phases and made participants aware that their compensation does not depend on their performance during the memory tasks at the beginning of the study.

\section{Ethical Considerations}
Although previous research has found generative AI less likely to change content and tone of the original message when paraphrasing in formal contexts (e.g. academic, news) than in informal contexts \cite{tripto-etal-2024-ship}, LLM revisions in this context can potentially introduce misinformation and inaccuracies. To prevent showing misinformation to participants during the study, we only included headlines which were judged as accurate by at least 62.5\% of annotators, with a mean accuracy acceptance rate of 80.6\% over all presented LLM revisions. 

For all data collected on Prolific, participants were compensated in GBP at a hourly rate equivalent to the current minimum wage in the country of this work (approx.~11 GBP). 
Participants who did not pass the qualification test were compensated for their time at the same rate. This corresponds to more than twice the federal minimum wage in the US, where all study participants and annotators were based. The student who rewrote the headlines for study III was employed at the university and compensated at an hourly rate in accordance to the official salary scale for research assistants in the country of this work.

All data was collected anonymously and does not allow conclusions about participants' identities. Participants in the memory study explicitly agreed to provided personal information (e.g. age, political orientation, summarized in Appendix \ref{sec:demographics}) being published in the informed consent before beginning the study.

\section*{Acknowledgements}

We thank Noas Shaalan for support in preparing the data for study III. We also thank the
anonymous CMCL reviewers for their valuable and constructive feedback.

\section{Bibliographical References}\label{sec:reference}

\bibliographystyle{lrec2026-natbib}
\bibliography{lrec2026-example}

\appendix
\section{Pilot Study}
\label{sec:pena}

\paragraph{Posthoc Analysis of \citeauthor{Pena2023-fw}'s data} We obtain \citeauthor{Pena2023-fw}'s study data and perform analyses using spacy's \cite{honnibal2020spacy} English transformer pipeline to identify linguistic features which may impact memorability of news headlines and tweets in their data. This post hoc analysis reveals significant correlations between memorability and pronoun use (Spearman Rank Coefficient $\rho$=0.31, p<0.001). At the same time, we also find that these characteristics are used significantly more in tweets than in news headlines (p<0.001) in the study items selected by \citeauthor{Pena2023-fw}. We were thus interested in whether these effects would persist, when only news headlines are used. In other words, we wondered if the increase in memorability of tweets compared to news headlines might stem from the language that is used, rather than the content or text type. To address this question, we performed a pilot user study using a within-subject design to reproduce findings from this analysis, using only naturalistic news headlines found in the wild and no tweets. 

\paragraph{Pilot Study} For our pilot study, we collected 32 news headlines from popular US news outlets with eight headlines for each of four topics: entertainment, politics, environment, and health. Within each topic, half the headlines contain pronouns and half do not. 60 participants were randomly split into two groups: each saw a different set of 16 headlines during the presentation phase, with the other 16 appearing first during the recognition phase. For both groups, headlines were balanced by topic and pronoun use. The study followed the same format as the main study. 

We conducted a mixed-effects linear regression to investigate whether the presence of pronouns in headlines influenced their recognition, taking into account participants’ experimental group. While neither the main effect of pronoun presence nor the main effect of experimental group reached statistical significance, their interaction did: headlines containing pronouns were recognized at different rates depending on the experimental group ($\beta = 0.73$, \textit{SE} = 0.31, \textit{z} = 2.39, \textit{p} = 0.017).

\begin{figure}[htbp]
    \centering
    \includegraphics[width=\linewidth]{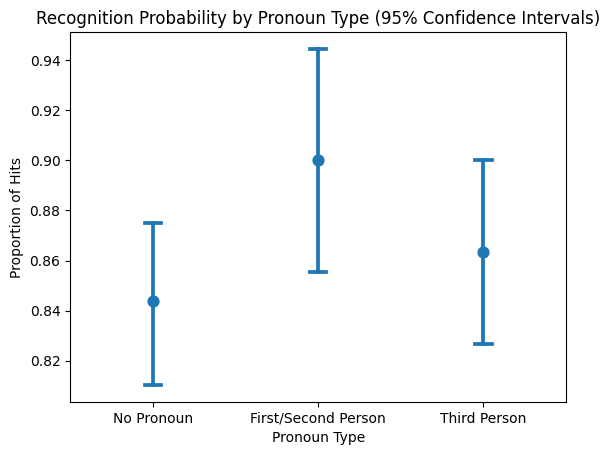}
    \caption{headline recognition likelihood given the contained pronoun type}
    \label{fig:pron_types}
\end{figure}

Follow-up analysis of individual headlines revealed that Group 2—where recognition rates for pronoun-containing headlines were higher—had a greater number of headlines using first- and second-person pronouns compared to Group 1. Across all participants, we also observed a consistent pattern: headlines featuring first- and second-person pronouns were more likely to be recognized than those without any pronouns, reinforcing the importance of pronoun type in headline recognition (see Figure \ref{fig:pron_types}).

\section{LLM Prompts and Setup}
\label{sec:models}
Our prompt is based on a combination of best-practice strategies suggested by \citet{white2023prompt}. We provide the LLM with a persona in the system prompt.

We also make use of the alternative approaches pattern and the reflection pattern described by \citet{white2023prompt}. After providing a description of the task and the constraints, LLMs are prompted to provide five revisions of a given headline. They are then asked to reflect on the quality of the headlines and choose the best headline based on specified criteria. The output is constrained by providing a response template and forcing JSON-format. If a model revision for a headline did not contain a first or second person pronoun on the first try (this happened for gpt-4o-mini, mistral, mixtral, and qwen), the generation process was repeated for the specific headline until the requirement was fulfilled.

\paragraph{System Prompt:}
   \textit{You are an editor at a high-quality newspaper. Your task is to subtly modify article headlines to make them more engaging, without altering the core message.\newline
Specifically, when given a headline, you will rewrite it by incorporating second-person (e.g., "you", "your") and/or first-person (e.g., "I", "my") pronouns. This will make the headline more relatable and attention-grabbing for the reader. Ensure that the revised headline remains true to the original tone and meaning. \newline
 Your goal is to make each headline more compelling and conversational, while maintaining clarity and relevance to the reader's experience. After generating multiple options, choose the one that fits best in terms of engagement, clarity, and relevance for the target audience.}

\paragraph{Prompt:}
\textit{You are given a news article headline. Your task is to rewrite it using first-person ("I", "my") and/or second-person ("you", "your") pronouns to make it more engaging and personally relevant to readers.\newline
Generate exactly five alternative versions of the headline. Each version should:\\
        - Preserve the original tone and core message as closely as possible.\\
        - Use first-person and/or second-person pronouns to create a direct, conversational appeal.\\
        After generating the five rewrites, analyze which one is the most effective. Your analysis should consider:\newline
        - Reader engagement\\
        - Clarity\\
        - Faithfulness to the original meaning\\
        Finally, select the single best version based on your reasoning.\newline
        Use the following json-format to return your output (no additional explanation or text):\newline
        $rewrite\_1$: First rewritten headline,\\
        $rewrite\_2$: Second rewritten headline,\\
        $rewrite\_3$: Third rewritten headline,\\
        $rewrite\_4$: Fourth rewritten headline,\\
        $rewrite\_5$: Fifth rewritten headline,\\
        $reasoning$: Explain why one version stands out in terms of engagement, clarity, and preservation of the original message.,\\
        $best\_headline$: The best headline from above\\
        Original headline: $``{row['headline']}$''}

\paragraph{Setup and Parameters} We used the default settings for all LLM calls, with the temperature parameter set to 0.3. GPT \cite{achiam2023gpt} and DeepSeek \cite{liu2024deepseek} models were accessed via the OpenAI API, while Mistral \cite{jiang2023mistral7b} and LLaMA \cite{grattafiori2024llama3herdmodels} ran on a single A40 GPU. Mixtral \cite{jiang2024mixtralexperts} and Qwen \cite{qwen3technicalreport} models were run on two A40 GPUs. Given the small number of headlines to process, generating all LLM outputs took less than an hour.

\section{Instructions for Human Annotation of Rewritten Headlines}
\label{sec:annotation}
The following instructions, adapted from the multidimensional quality metrics framework \citet{burchardt-2013-multidimensional} were provided to annotators for the evaluation of LLM-revised headlines. The instructions were followed by a list of 6 more examples. Headlines used for examples and the qualification test did not appear in the main annotation task.
\newline

\noindent{\itshape \textbf{Overview}\newline
You will see two versions of a news headline, one marked as "Original", the other as "Revision". A revision is a rewriting  of the original headline, so that it includes one or more first or second person pronouns. Your task is to annotate whether the revised version retains the orsiginal content and style and to what extent it reflects a news headline you are likely to read on typical news outlets. \newline
You are asked to judge the following categories:\newline
\textbf{1. Accuracy: Does the content in the revised version accurately reflect the content of the source text?  }\newline
There are three ways, in which accuracy is commonly violated, which can also occur together: 
\begin{itemize}
    \item \textbf{Misrepresentation}: The revision misrepresents information provided in the original headline       
\item \textbf{Addition}: The revision includes content not present in the original headline        
\item \textbf{Omission}: The revision is missing content present in the original headline
\end{itemize}
\textbf{2. Style: Is the language style of the revised headline appropriate?}    
Inappropriate style can manifest in various forms, that can also occur together: 
\begin{itemize}
    \item \textbf{Grammar}: The revision contains grammar or language errors   
\item\textbf{Awkward Style}: The revision is grammatical, but unnatural as a news headline or awkward (e.g it involves excessive wordiness or overly embedded clauses)    
\item\textbf{Inconsistent Style}: The style or tone is inconsistent within the revision (e.g. factual, dry information is paired with sensationalism)
\end{itemize}

\noindent If a revision differs in tone or emotion compared to the original version, you can indicate this in a separate checkbox.
Also, feel free to add comments in the comment field. You can even suggest revision improvements, if you can think of a better phrasing (but this is not the main goal of this annotation task)\\\\
\textbf{Examples:}\\\\
\textbf{Original:} Jonathan Majors reportedly admits to being ‘aggressive’ with ex-girlfriend in newly released audio clip \newline
\textbf{Revision: }You need to hear Jonathan Majors' disturbing admission about being 'aggressive' with an ex-girlfriend   \newline
1. Does the content in the revised version accurately reflect the content of the source text?\newline   
[x] Yes     [ ] No\newline
2. Is the language style of the revised headline appropriate?  \newline
[x] Yes     [ ] No\newline
[x] The revision differs in tone or emotion compared to the original\\
\textbf{Explanation}: The headline is appropriate as a news headline and accurately reflects the content of the original headline. The addition of the word ``disturbing'' changes the emotional tone of the revision compared to the original.}

\section{Instructions for Human Revisions}
\label{sec:revisions}
The following instructions were given to participants when collecting human revisions of news headlines. The same instructions were also given to the graduate student who revised headlines for study III.\newline

\textit{You are given a set of news article headlines, one after the other. Your task is to rewrite each headline using first-person (e.g. "I", "my", "our") and/or second-person (e.g. "you", "your") pronouns to make it more engaging and personally relevant to readers.\newline
Your revision should preserve the original tone and core message as closely as possible and use one or more first-person and/or second-person pronouns to create a direct, conversational appeal.\newline
For some headlines, this might be easier than for others. Feel free to make changes to the structure or wording of a headline if needed, but make sure the content and tone stay faithful to the original headline. }

\section{Demographics of Memory Study Participants}
\label{sec:demographics}
\paragraph{Study I} participants were between 20 and 74 years old (mean age: 42.41). 29 identified as female and 28 as male and 3 chose not to disclose. 56.14\% held a Bachelor's or Master's degree, 21.05\% had some college education, 14.04\% had only a high school degree and the rest held associate or professional degrees. The majority of participants (64.91\%) were employed. When asked about their political views on a five point likert scale ranging from 1 - left  to 5 - right, 19.3\% indicated political affiliation with the left and 7.02\% with the right, whereas the rest fell between. 28.07\% of participants indicated they consumed news on news websites more than once a day and 28.07\% once a week or less, with the rest falling in-between. 55.36\% consumed news on social media more than once a day, and 19.65\% once a week or less.

\paragraph{Study II} participants were between 22 and 67 years old (mean age: 41.98). 33 identified as female, 21 male, 2 non-binary and 4 chose not to disclose. 60.72\% held a Bachelor's or Master's degree, 21.43\% had some college education, 8.93\% had only a high school degree and the rest held associate or professional degrees. The majority of participants (67.86\%) were employed. 36.36\% indicated political affiliation with the left and 23.64\% with the right, whereas the rest fell between. 34.55\% of participants indicated they consumed news on news websites more than once a day and 14.55\% once a week or less, with the rest falling in-between. 66.07\% consumed news on social media more than once a day, and 12.5\% once a week or less.

\paragraph{Study III} participants were between 21 and 68 years old (mean age: 40.23). 54 identified as female, 66 male. 65\% held a Bachelor's or Master's degree, 13.33\% had some college education, 10\% had only a high school degree and the rest held associate or professional degrees. The majority of participants (78.33\%) were employed. 25.83\% indicated political affiliation with the left and 25\% with the right, whereas the rest fell between. 26.89\% of participants indicated they consumed news on news websites more than once a day and 25.21\% once a week or less, with the rest falling in-between. 65\% consumed news on social media more than once a day, and 10\% once a week or less. 
\end{document}